# An attempt to generate new bridge types from latent space of PixelCNN

Hongjun Zhang

Wanshi Antecedence Digital Intelligence Traffic Technology Co., Ltd, Nanjing, 210016, China

583304953@QQ.com

**Abstract:** Try to generate new bridge types using generative artificial intelligence technology. Using symmetric structured image dataset of three-span beam bridge, arch bridge, cable-stayed bridge and suspension bridge, based on Python programming language, TensorFlow and Keras deep learning platform framework, PixelCNN is constructed and trained. The model can capture the statistical structure of the images and calculate the probability distribution of the next pixel when the previous pixels are given. From the obtained latent space sampling, new bridge types different from the training dataset can be generated. PixelCNN can organically combine different structural components on the basis of human original bridge types, creating new bridge types that have a certain degree of human original ability. Autoregressive models cannot understand the meaning of the sequence, while multimodal models combine regression and autoregressive models to understand the sequence. Multimodal models should be the way to achieve artificial general intelligence in the future.

## 0  Introduction

In the long river of human history, the emergence of many new technologies such as materials, mechanics, computers, etc. has had a huge or even revolutionary impact on bridge engineering. Before the 17th century, bridges were generally constructed using wood and stone materials, which were limited by the mechanical properties of the materials and had a maximum span of only a few tens of meters. After the Industrial Revolution in the 18th century, the production of iron, steel, and concrete provided new construction materials for bridges, and soon the maximum span of bridges could reach several hundred meters. Under the guidance of deflection theory, the span of the Golden Gate Bridge completed in 1937, reached 1280 meters. The computer-based finite element analysis method, which emerged in the 1960s, can accurately analyze the mechanical behavior of complex structures, enabling the realization of modern cable-stayed bridge types. At present, the maximum span of the bridge has exceeded 2000 meters, which cannot be achieved without advanced materials, mechanical theories, and calculation methods.

The current artificial intelligence technology has provided new impetus for the development of civil engineering. It is believed that the era of intelligent construction is approaching [1], and the bridge type innovation in this article is only a drop in the ocean of its application. It can be imagined that the near future: virtual assistants use a mixed reality approach to report their creations to human engineers; Intelligent machines are busy at the villa construction site, while humans are drinking coffee and longing for the day they will be built and moved in.

The author's previous papers [2-3] trained Variational Autoencoder (VAE) and Generative Adversarial Network (GAN), successfully generating new bridge types that are completely different from the dataset, and concluded that they can assist bridge designers in bridge type innovation. Among numerous generative artificial intelligence technologies, Autoregressive Models, Energy-Based Models, and Diffusion Models, etc. can also be applied to bridge innovation.

Pixel Convolutional Neural Network (PixelCNN) imagines an image as a sequence of pixels, calculates the probability distribution of the next pixel based on previous pixels, and generates an image pixel by pixel. This article uses PixelCNN, based on the same dataset as before, to further attempt bridge innovation (open source address of this article's dataset and source code: https://github.com/QQ583304953/Bridge-PixelCNN).

# 1 Introduction to PixelCNN

## 1.1 Overview of Autoregressive Models

Regression models estimate the dependent variable (Y) based on the independent variable (X), for example, predicting stock market trends based on fundamental analysis (positive or negative concepts).

Autoregressive models estimate current values based on their own past values, for example, predicting current stock market trends solely on the stock market's own historical data (Candlestick Charts technique analysis).

Autoregressive models can reveal the inherent patterns and trends of time series data, and common models include LSTM, PixelCNN, and GPT. It has defeated human chess players and can profit from trend-following strategy in the stock market. It can predict earthquakes by analyzing historical records of earthquakes, making ChatGPT chat and communicate like humans.

## 1.2 Steps for generating texts character by character of LSTM

Taking section 8.1 of reference 《Python Deep Learning》[4] P227 as an example: ① First, extract a large number of samples and targets from the corpus (the sample is a 60 characters sequence, with the target being the next character of the sample); ② Train LSTM language model; ③ Then provide the initial text (seed sequence), input it into the model, and obtain the probability distribution of the next character. Randomly sample according to the probability value (if the probability of a character is 0.3, there is a 30% probability of selecting it), and continuously generate a whole paragraph of text character by character, as shown in Figure 1.

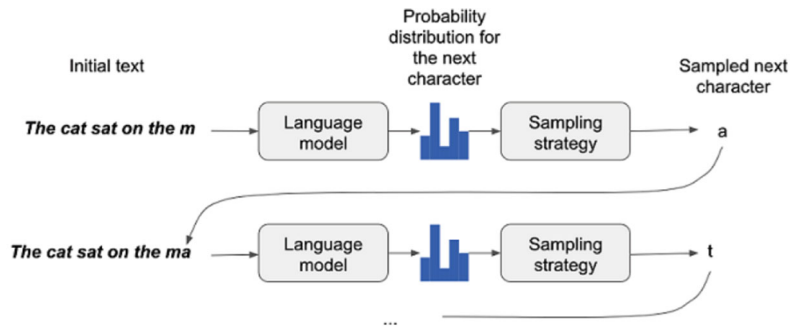

Fig.1 The process of character-by-character text generation using a language model

## 1.3 Steps for generating images pixel by pixel of PixelCNN

The process is exactly the same as LSTM's character-by-character text generation, taking grayscale images as an example:

(1) Imagine the image as a sequence of pixels[5-7], and index number of each pixel of the image with a number from left to right and from top to bottom.

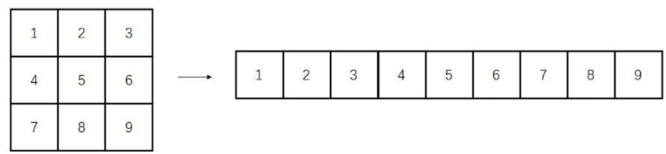

Fig.2 The image is converted to a sequence of pixels

(2) Dataset: A small area of the image is used as a sample, and the target is the central pixel value of that area. When using a 3x3 convolution kernel, pixel numbers 1-4 in Figure 2 represent sample, pixel numbers 5 represent target, and pixel numbers 6-9 are masked as they are not related to the task. Each 192x48 pixel grayscale image is equivalent to 192*48=9216 training samples.

(3) Train PixelCNN。

(4) The parameter padding of the Conv2D layer is "same", so that edge pixels can also serve as the center of the convolution window, which is equivalent to giving a sequence of seed pixels. Then, the model outputs the probability distribution of the next pixel based on the previous pixels, randomly samples according to the probability value, and generates an image pixel by pixel.

When generating an image, first set an empty image, and then the model code performs calculations from beginning to end to calculate the predicted value of the first pixel; Then fill the predicted value of the first pixel into the empty image, input it into the model, and obtain the second pixel, ……. The model runs as many times as there are pixels in an image, and different pixels in a single image cannot be calculated in parallel, so the generation speed is very slow. This results in the need for high computational power to monitor the real-time image generation effect, and model debugging takes a long time.

### 1.4 Latent space of PixelCNN

Each dimension of the sample space of the images is highly correlated (such as the values of pixels far apart in a symmetric image are equal). The number of spatial dimensions equals the number of pixels, and the number of discrete value of each dimension equals the number of pixel categories. The sample space is very large, with space capacity=number of pixel categories to the power of number of dimensions, which is an astronomical number, and most of the sample points in space have no practical significance.

How to handle the dependency relationships between spatial dimensions and how to find meaningful samples from space are the challenges of all algorithms of generative deep learning. Directly modeling the joint probability distribution in the sample space may be difficult to implement due to the large number of required samples and model parameters.

If the naive Bayesian algorithm is used for modeling (number of dimensions in the latent space = number of dimensions in the original sample space, and assuming that each dimension in the latent space is completely independent), it will completely separate the relationships between each dimension. This leads to a huge deviation from the actual situation, and the possibility of obtaining satisfactory samples by sampling in such a high-dimensional latent space is like a monkey continuously typing a keyboard to write a novel.

Unlike the low-dimensional compression and dimensionally independent latent space of VAE and GAN, the latent space of PixelCNN maintains the same number of dimensions of the original sample space. It captures the dependency relationship between adjacent dimensions in the latent space through data statistics, and approximates the joint probability distribution by the product of conditional probabilities. Specifically, ①The joint probability distribution of the images in the original sample space is $p(x)= p(x_1, x_2, \cdots x_n)$; ② Establish a latent space with the same number of dimensions as the original sample space, set a pixel sequence (order of dimensions), assuming that the value of a pixel only depends on its previous pixel values, that is, $p(x_i)= p(x_{i-1}) * p(x_i | x_{i-1})$; ③ In this way, the joint probability distribution can be approximated as the product of conditional probabilities [5-7], that is $p(x)= p(x_1)* p(x_2 | x_1)* p(x_3 | x_2) \cdots p(x_n | x_{n-1})$.

Obviously, this approximation operation of PixelCNN is not precise enough and may have some deviation from the actual situation, but the key is to simplify complex problems and facilitate statistical learning. This type of autoregressive model has achieved good application results in fields such as text generation and speech generation.

## 2 An attempt to generate new bridge types from latent space of PixelCNN
### 2.1 Dataset

Using the dataset from the author's previous paper [2-3], which includes two subcategories for each type of bridge (namely equal cross-section beam bridge, V-shaped pier rigid frame beam bridge, top-bearing arch bridge, bottom-bearing arch bridge, harp cable-stayed bridge, fan cable-stayed bridge, vertical_sling suspension bridge, and diagonal_sling suspension bridge), and all are three spans (beam bridge is 80+140+80m, while other bridge types are 67+166+67m), and are structurally symmetrical.

This model has a large number of parameters and a very slow image generation speed, so the image size is reduced from 512x128 to 192x48 pixels (the disadvantage is that the clarity is much reduced).

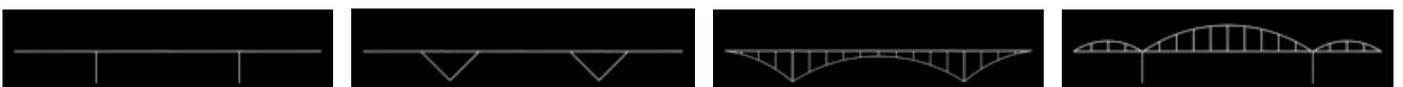

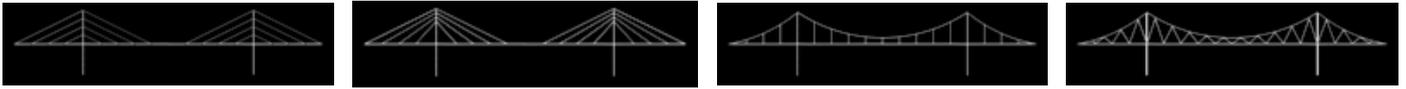

Fig.3 Grayscale image of each bridge facade

Each sub bridge type obtained 1200 different images, resulting in a total of 9600 images in the entire dataset.

## 2.2 Construction of PixelCNN

Based on the Python3.10 programming language, TensorFlow2.10, and Keras2.10 deep learning platform framework, construct and train PixelCNN. The overall architecture diagram is as follows:

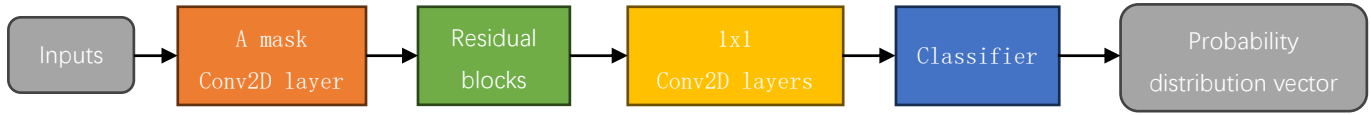

Fig.4 Overall architecture of pixelcnn

(1) A mask Conv2D layer

To ensure that the output of each pixel layer is only affected by the pixel values before the relevant pixels, it is necessary to mask the local area of the convolutional filter window. This is achieved by multiplying the mask with the filter weight matrix, in order to reset the value of any pixel after the target pixel to zero[5-7].

The initial masked convolutional layer cannot use the central pixel because it is exactly the pixel we want the network to guess, that is, using A mask.

Subsequent layers (residual blocks, 1x1 Conv2D layers) can use the central pixel, as it is an intermediate result calculated based on the information of previous pixels in the original input image, that is, using B mask.

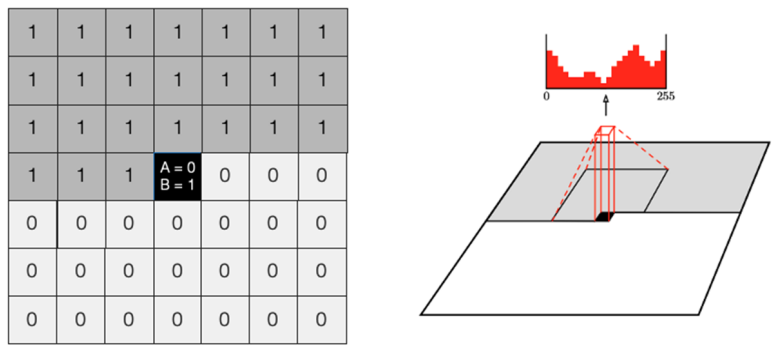

Fig.5 Left-convolution filter mask, type A covers the center pixel, type B does not cover the center pixel. Right – A mask applied to a set of pixels to predict the distribution of central pixel values

(2) Residual block

A residual block is a set of layers where the output is added to the input before being passed on to the rest of the network. In other words, the input has a fast-track route to the output, without having to go through the intermediate layers—this is called a skip connection.

Residual connections can solve the problem of vanishing gradients, and representational bottlenecks[4].

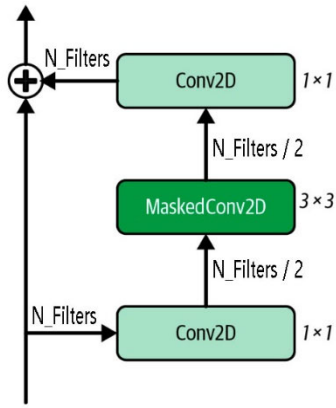

Fig.6 A residual block in PixelCNN

(3) 1x1 Conv2D layers

Pointwise convolution operation becomes equivalent to running each pixel vector(multiple channel values) through a Dense layer. It will compute features that mix together information from the channels of the input tensor, but it will not mix information across space at all (since it is looking at one pixel at a time).It helps to distinguish channel feature learning from spatial feature learning. If each channel is highly autocorrelated when crossing space, but different channels may not be highly correlated, then this approach is reasonable[4].

(4) Classifier

The output layer is a softmax layer that predicts all possible categories of pixels.

## 2.3 Training

(1) The initial version of PixelCNN has a challenge, which is that the model can not understand that pixel values 254 and 255 are very close, and it has to independently learn each pixel output value. Too many target categories results in a significant increase in the number of training epochs required.

Assuming there are 2 pixel samples, targets are [254, 254]. The probability distribution of 256 pixel values predicted by PixelCNN is [[0.01, 0.06, ···, 0.8, 0.05], [0.02, 0.03, ···, 0.05, 0.8]], which means that the maximum probability predicted by the first sample corresponds to a pixel value of 254, and the maximum probability predicted by the second sample corresponds to a pixel value of 255. From the perspective of human visual perception, the predictions are both very accurate. And cross entropy (sparse_categorical_crossentropy) is [0.22, 3.0], and there is a huge difference in the loss between the two samples. In the model's view, this is an error in predicting the category.

One of the improved methods is to reduce the number of categories, dividing the 0-255 pixel values into several intervals (an interval is a categoriy), so that each pixel can only take one of several categories. The decrease in the number of categories makes the model easier to solve, but the cost is that the image can only be represented by a few colors (this problem can be very prominent in color images).

(2) When the probability distribution curve shows continuous and smooth, the difference in probability between adjacent pixels will be small, which means that the model can understand that pixel values 254 and 255 are very close, rather than differences in categories. This is the main improvement of PixelCNN++version [8], which uses logistic distribution to calculate probability. Discretize the logistic distribution into 256 intervals, each interval corresponding to 0-255 pixel values. The model outputs two parameters, mean value and standard deviation, which can completely control the height and weight of the logistic curve, so as to accurately control the probability of each interval.

Here, we directly use the PixelCNN class (PixelCNN++version) from the TensorFlow Probability library [9], which has superior performance. By directly applying the MNIST instance code on the official website, we can achieve preliminary results, greatly facilitating users. Fine-tuning the four parameters ( num_resnet, num_hierarchies, num_logistic_mix, receptive_field_dims) can get a satisfactory result.The specific parameters are as follows:

```python
dist = tfp.distributions.PixelCNN(
    image_shape=(48, 192, 1),
    num_resnet=3, #the number of ResNet layers within each highest-level block
    num_hierarchies=1, #the number of hightest-level blocks (in Figure 2 of arXiv:1701.05517v1)
#View Source "pixel_cnn.py": "for i in range(self._num_hierarchies)". Each `i` iteration builds one of the
#highest-level blocks (identified as 'Sequence of 6 layers' in the figure 2 of arXiv:1701.05517v1, consisting
#of `num_resnet=5` stride-1 layers, and one stride-2 layer that contracts the height/width dimensions).
    num_filters=32, #the number of convolutional filters
    num_logistic_mix=1, #number of components in the logistic mixture distribution.
#Taking 1 when grayscale images can speed up the loss reduction
    receptive_field_dims=(5,7), #height and width in pixels of the receptive field of the convolutional layers
#View Source "pixel_cnn.py": rows, cols = receptive_field_dims;  Conv2D( kernel_size=(2 * rows - 1, cols),...)
#So,the first value is the maximum number of rows of visible pixels,
#and the second value is the number of convolution kernel window columns
    dropout_p=0.3,
) # https://tensorflow.google.cn/probability/api_docs/python/tfp/distributions/PixelCNN
```

Fig.7 Parameter Settings of pixelcnn class in tensorFlow probability library

Due to the constraints of hardware conditions, it is impossible to fully optimize. Thus, I can only test parameters while sampling, and generate bridge images from multiple saved generator models.

## 2.4 Exploring new bridge types through latent space sampling

Random sampling. Based on the thinking of engineering structure, twelve technically feasible new bridge types are obtained through manual screening, which were completely different from the dataset (Figure 8, black character on white background).

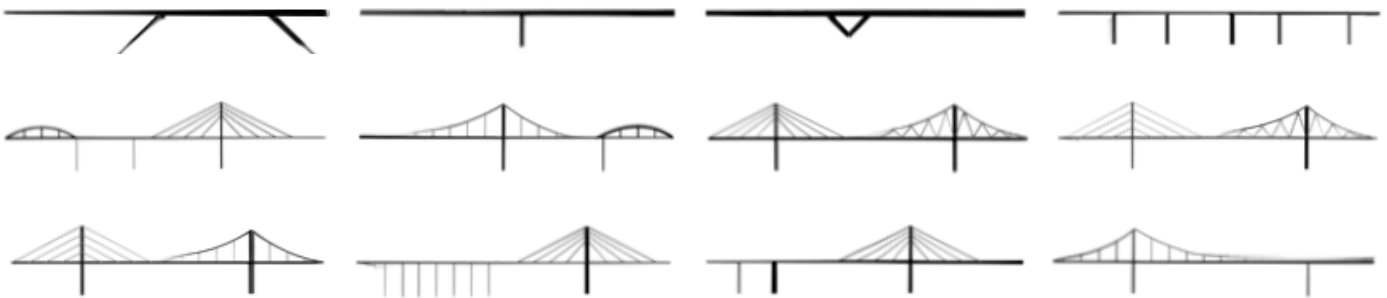

Fig.8 Twelve new bridge types with feasible technology

The new bridge type here refers to a type of bridge that has never appeared in the dataset, but is created by neural network based on algorithms, which represents the model's innovative ability. For example, in the first image of the above figure, "Slant legged rigid frame bridge " (in reality, it is a subclass of beam bridge). Some bridge types, such as cable-stayed bridge with single pylon, are very common in reality.

The longitudinal combination of cable-stayed bridge and suspension bridge, to my knowledge, has not been practiced by any engineer due to complex stress or unreasonable structure. When designing ultra large span bridges, human engineers strive to simplify the structure as much as possible. Therefore, this combination bridge type is not suitable for ultra large span bridges, but can be used for urban landscape bridges.

## 2.5 Result analysis

The bridge types of the dataset are all symmetric structures, while PixelCNN can generate asymmetric bridge types, which are not simply superposition, but organic combinations of different structural components. It is similar to generative adversarial networks.

## 3 Multimodal models are the way to achieve artificial general intelligence

The disadvantage of autoregressive models is obvious. They only analyze their own historical datas and determine the next value based on probability distribution. The models do not understand the meaning expressed by the sequences. Just like learning a foreign language solely through audio (without any guidance or language scenes), it can only reach the level of Parrot Talk.

It is the information outside the sample space that determines the meaning of the samples in space, and the samples in space are the result rather than the cause. Autoregressive models only perform statistical learning on samples, treating the symptoms rather than the root cause. Therefore, more advanced models in the future need to obtain information beyond the sequence.

Human intelligence has the characteristics of both regression and autoregressive models. In life, people organize language and actions based on realistic scenes, which is a regression model behavior. In 1945, Mr. Huang Yanpei's famous thinking about the "cyclical law of Chinese history" (referring to the cyclical phenomenon of the rise and fall of the political power in Chinese history) is based on the behavior of the autoregressive model.

Multimodal models are comprehensive imitation of human intelligence, combining regression and autoregressive models, similar to learning a foreign language through matching audio with scenes, in order to grasp the meaning of audio and overcome the shortcomings of autoregressive models. I personally believe that multimodal models should be the path to achieving artificial general intelligence in the future.

## 4 Conclusion

(1) PixelCNN is similar to generative adversarial network and is more creative than variational autoencoder. It can organically combine different structural components on the basis of human original bridge types, creating new bridge types. It has a certain degree of human original ability, and can open up the imagination space and provide inspiration to humans.

(2) The disadvantage of PixelCNN is that the sampling speed is slow, and training and actual deployment require high computational power.

(3) Autoregressive models cannot understand the meaning expressed by sequences, while multimodal models combine regression and autoregressive models to understand sequences. Multimodal models should be the path to achieving artificial general intelligence in the future.